%% file: emnlp_main.tex
\pdfoutput=1

\documentclass[letterpaper, 11pt]{article}

\usepackage[]{conf/acl}

\usepackage{times}
\usepackage{latexsym}
\usepackage{breqn}

\usepackage[T1]{fontenc}

\usepackage[utf8]{inputenc}

\usepackage{microtype}
\usepackage[ruled,vlined]{algorithm2e}
\usepackage{graphics}
\usepackage{booktabs}
\usepackage{tabularx}
\usepackage{tablefootnote}
\usepackage[export]{adjustbox}
\usepackage{subcaption}
\usepackage[font=small]{caption}

\usepackage{graphicx}
\usepackage{xcolor}

\title{Generative Aspect-Based Sentiment Analysis \\with Contrastive Learning and Expressive Structure}

\author{Joseph J. Peper \and Lu Wang \\
  Computer Science and Engineering \\
  University of Michigan \\
  Ann Arbor, MI \\
  \texttt{\{jpeper, wangluxy\}@umich.edu} \\}

\begin{document}
\maketitle
\begin{abstract}

Generative models have demonstrated impressive results on Aspect-based Sentiment Analysis (ABSA) tasks, particularly for the emerging task of extracting Aspect-Category-Opinion-Sentiment (ACOS) quadruples. However, these models struggle with \textit{implicit} sentiment expressions, which are commonly observed in opinionated content such as online reviews.
In this work, we introduce \textbf{GEN-SCL-NAT}, which consists of two techniques for improved structured generation for ACOS quadruple extraction.
First, we propose \textbf{GEN-SCL}, a supervised contrastive learning objective that aids quadruple prediction by encouraging the model to produce input representations that are discriminable across key input attributes, such as sentiment polarity and the existence of implicit opinions and aspects. Second, we introduce \textbf{GEN-NAT}, a new structured generation format that better adapts autoregressive encoder-decoder models to extract quadruples in a generative fashion. \\
Experimental results show that GEN-SCL-NAT achieves top performance across three ACOS datasets, averaging 1.48\% F1 improvement, with a maximum 1.73\% increase on the LAPTOP-L1 dataset. Additionally, we see significant gains on implicit aspect and opinion splits that have been shown as challenging for existing ACOS approaches.

\end{abstract}

\begin{figure*}[t]
\centering
     \includegraphics[width=0.9\textwidth]{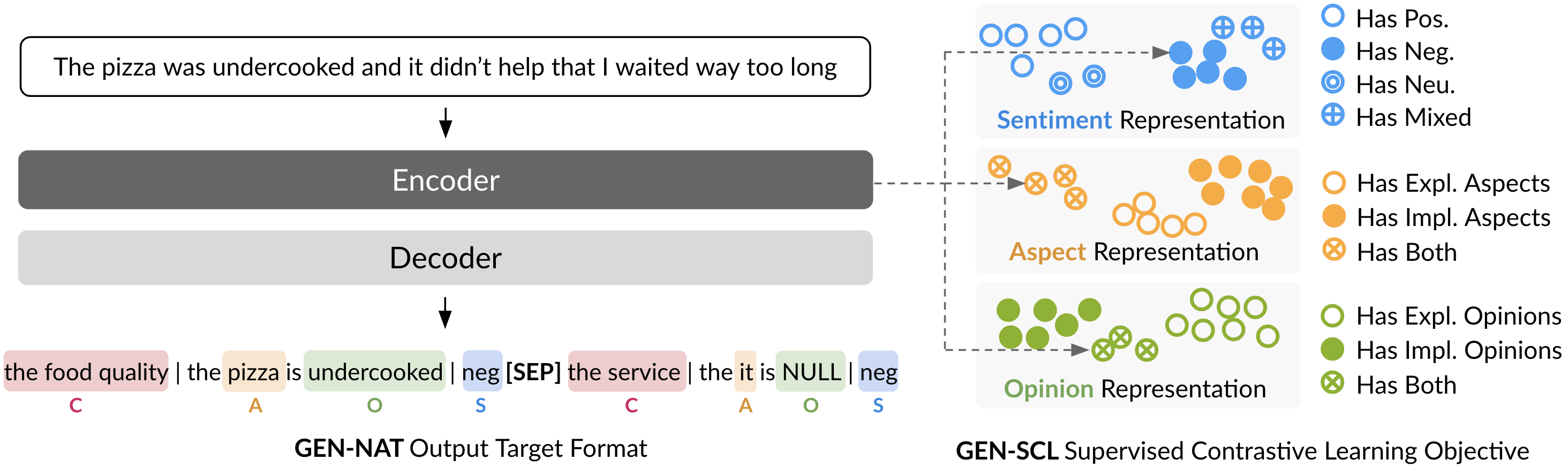}
      \caption{Overview of the GEN-SCL-NAT ACOS approach. GEN-NAT uses natural category descriptions and intuitive quadruple ordering to align the target output with the encoder-decoder's pre-training and auto-regressive nature. GEN-SCL trains the model to discriminably represent three sentiment characteristics during training: (1) sentiment polarity, (2) aspect term type, and (3) opinion term type. For each characteristic, we project the summed encoder final hidden states into a 1024-dimension characteristic representation via a simple fully-connected layer.
}
     \label{fig:task_overview}
     \vspace{-7pt}
\end{figure*}

\section{Introduction}

Aspect-based sentiment analysis (ABSA) is the task of extracting fine-grained sentiment information from text. Applications of this task range from social media opinion mining and empathetic dialog assistants to product review analysis \cite{surveyzhang2022}. 

Common ABSA subtasks involve identifying \textit{aspect term} mentions and their corresponding aspect \textit{categories}, associating them with supporting \textit{opinion terms}, and/or \textit{sentiment polarity} that is implicitly or explicitly expressed within the text.

While over 30\% of sentiment expressions contain implicit language \cite{acos}, existing methods struggle with these, performing significantly worse on examples with implicit aspects and/or opinions. These cases, such as ``it took an hour to be seated'', are more difficult for models as they lack common indicative \textit{explicit} aspect and opinion terms, e.g., ``the service'' and ``slow''. 

In this work, we address the ACOS quadruple extraction task, particularly for formulations supporting implicit aspects and opinions. We build off \citet{asqp}, a state-of-the-art T5-based technique outputting parseable structured ACOS quadruple predictions for a given text. We introduce \textbf{GEN-SCL-NAT}\footnote{Code and models are made available at \url{https://github.com/jpeper/GEN_SCL_NAT}.}, 
consisting of two novel modifications to this approach. First, we modify the model training objective, adding \textbf{GEN-SCL}, an auxiliary \underline{s}upervised \underline{c}ontrastive \underline{l}earning objective that tasks the model to discriminably represent examples across key characteristics such as the existence of implicit aspects and opinions, and the expressed sentiment polarity. A supervised contrastive loss is used to maximize the margin between inconsistent examples and minimize it between consistent examples. 
Second, we introduce a modified generation target format with \textbf{GEN-NAT}, where we improve the \underline{nat}uralness of the target format in three ways: by (1) replacing existing aspect category labels with human-readable category descriptions, (2) enforcing a reproducible ``scan-based'' inter-quadruple ordering for multi-quad cases, and (3) modifying the intra-quadruple format to reflect a natural causal ordering of quadruple components that aligns with auto-regressive decoding.
Figure \ref{fig:task_overview} outlines the ACOS task and our GEN-SCL-NAT approach.

Leveraging these techniques, we show improved performance on ACOS benchmarks, with respective ACOS F1 improvements of 1.64\%, 1.08\% and 1.73\% on the REST, LAPTOP, and LAPTOP-L1 datasets. We also see significant gains on examples with challenging implicit language, averaging 1.47\% improvement on this subset. Finally, we conduct an ablation study that shows the complementary behavior of GEN-NAT and GEN-SCL.

\section{ACOS Quadruple Extraction Task Formulation}
We follow \citet{asqp} and \citet{acos} in formulating ABSA as a joint quadruple extraction task, where the goal is to extract an unordered set of ACOS quadruples $Q_1, Q_2, ... Q_n$ in text $T$, where $Q_i$ = \((a_i, c_i, o_i, s_i)\) contains a corresponding aspect term, aspect category, opinion term, and sentiment polarity. Quadruples can lack clear supporting aspect and/or opinion spans, and these cases are marked as \textit{implicit}. Figure \ref{fig:task_overview} displays the ACOS task and quadruple components.

\section{Methodology}
In this section we (1) introduce \textit{GEN-SCL}, a task-specific supervised contrastive learning (SCL) objective, and (2) propose \textit{GEN-NAT}, an enhanced ACOS quadruple generation target format that addresses weaknesses in existing methods.

\subsection{GEN-SCL Supervised Contrastive Loss}

We propose GEN-SCL, an auxiliary SCL objective that encourages the encoder-decoder model to discriminably represent several key characteristics of the input while concurrently fine-tuning for the downstream generation task of ACOS quadruple extraction.
We task the model with learning representations of example-level \textbf{Sentiment}, \textbf{Aspect} and \textbf{Opinion}  characteristics. Figure \ref{fig:task_overview} indicates the label sets for each characteristic.

\paragraph{Representation Generation Process}Similar to \citet{scaptABSA}, we generate representations for example $x_i$ by feeding the sum-pooled encoder representation $Mean(Encode(x_i))$ through a single unique fully-connected layer $FC_c$ for each characteristic $c \in {\{Sentiment, Aspect, Opinion\}}$. This generates representation $\textbf{h}_{ci}$. In our experiments the dimensionality of the input and outputs to the fully-connected layers are both 1024.

\paragraph{SCL Formulation}
Supervised contrastive learning encourages the model to maximize the representation similarity between \textit{same-label} examples, and to minimize it for \textit{different-label} examples. We follow \citet{supcl-seq} in their general SCL formulation, where for characteristic $c$ and training example $x_i$ in training mini-batch $M$, the loss is: 
\begin{equation}
\small
\mathcal{L}_{i}^{c}=\frac{-1}{|P(i)|}
\sum_{p \in P(i)} \log \frac{e^{\operatorname{sim}\left({\mathbf{h}}_{ci}, {\mathbf{h}}_{cp}\right) / \tau}}{\sum_{b \in B(i)} e^{\operatorname{sim}\left({\mathbf{h}}_{ci}, {\mathbf{h}}_{cb}\right) / \tau}}
\end{equation}

To ensure each example $x_i \in M$ has at least one \textit{same-label} example for comparison, we first extend $M$ with one dropout-altered view per mini-batch element, perturbing each example representation $\mathbf{{h}_{ci}}$ with dropout probability $p=0.1$ while maintaining the original label. This becomes $M^{*}$. 
$B(i) \equiv M^{*}\setminus{x_i}$ is all other examples in the extended mini-batch, and $P(i) \equiv \left\{p \in B(i): {y}_{cp}={y}_{ci}\right\}$ is the subset with a matching label.

\paragraph{Final training objective} We add to the existing decoder cross-entropy loss $\mathcal{L}_{C E}$ our three characteristic-specific losses:
 
\begin{dmath}
\mathcal{L}_{total}=\mathcal{L}_{CE} +\alpha_1\mathcal{L}_{SCL\_sentiment} +\alpha_2\mathcal{L}_{SCL\_aspect} +\alpha_3\mathcal{L}_{SCL\_opinion}
\end{dmath}

In our experiments we set $\alpha_1 = \alpha_2 = \alpha_3 = \alpha$, tuning $\alpha$ and the SCL temperature $\tau$ on the dev set, with values reported in Table \ref{tab:scl_params}.

\subsection{GEN-NAT Structured Generation Format}
\paragraph{Existing Structured Generation Formulation}

We build off \citet{asqp} who define the generation target as a linearization of $Q_i = (a_i, c_i, o_i, s_i)$ to an output format $P$:

\begin{equation}
\small
    P_c(c_i) \, is\, P_s(s_i) \,because\, P_a(a_i) \,is \,P_o(o_i)
\end{equation}
$P_c(c_i) = c_i$ and $P_s(s_i)$ is the mapping of $s_i$ options [positive, neutral, negative] to [\textit{``great'', ``okay'', ``bad''}]. $P_a(a_i) = a_i$ and $P_o(o_i) = o_i$ for explicit cases, but are respectively ``it'' and ``null'' for implicit cases.
Quadruples are concatenated with a separator token to form the final output:
\begin{equation}
\small
    P(Q_1) \; [SSEP] \; ... \; [SSEP] \; P(Q_n)
\end{equation}

\paragraph{GEN-NAT Structured Generation Modifications}
\label{GEN-NAT}
We implement three modifications to the existing paraphrase generation format to aid in decoding.\\
(1) We revise $P_c(c_i)$, now defining it as a mapping of the raw $c_i$ (e.g. ``Laptop\#Usability'') to a natural category description (e.g. ``the laptop usability''). Note: we also change the sentiment linearization $P_s(s_i)$, simply using [\textit{``positive'', ``neutral'', ``negative''}], as we observe \textit{``okay''} can often imply negative sentiment within review text (e.g ``The food was just okay. I wouldn't return.'').
(2) For multi-output cases, per \citet{keyphrase-gen}, a consistent ordering is useful in training even when predicting unordered sets; following them, we use a ``scan-based'' ordering, outputting quadruples by their last-occurring explicit aspect or opinion term. Quadruples with only implicit aspect and opinion spans are generated last in random order. 
(3) We revise the quadruple linearization format, following \citet{seq2path} in outputting the quadruple elements in a natural top-down causal ordering: (c, a, o, s). Now, sentiment prediction is conditioned on aspect and opinion outputs. We also partition some quadruple components with ``|'' to reduce parsing ambiguity when mapping the predictions back to ACOS format.

Our resultant NAT generation format is as follows:
 \begin{equation}
 \small
    P_c(c_i) \;| \; the \;P_a(a_i) \;is \;P_o(o_i) \; | \; P_s(s_i) \
\end{equation}

\section{Experiment Setup}
We detail the experiment setup for evaluating our techniques on the ACOS task.

\begin{table}[t]
\centering
    \include{tables/dataset_stats}
    \caption{
    Dataset statistics. Over 33\% of REST quadruples contain implicit language, as do 43\% in LAPTOP*. E: explicit, I: implicit, A: aspect, O: opinion. E.g., \textit{IAEO} refers to ``implicit aspect, explicit opinion''. 
    }
    \label{tab:dataset}
    \vspace{-3mm}
\end{table}

\paragraph{ACOS Datasets}
Table \ref{tab:dataset} reports dataset statistics. The RESTaurant and LAPTOP datasets \cite{acos} are drawn from restaurant and e-commerce domains and are ACOS-labeled reviews including implicit aspects and opinions. 
LAPTOP-L1 differs from LAPTOP only by the category label granularity, using only the 21 top-level categories of the two-level category hierarchy in LAPTOP.

\begin{table}[t]
\centering
    \include{tables/primary_results}
    \caption{Overall F1 performance of quadruple extraction techniques on the REST, LAPTOP and LAPTOP-L1 datasets. Scores are averaged over 5 unique runs. $^*$Results are from \citet{acos} and \citet{seq2path}. $\dagger$: method is significantly better than PARAPHRASE (one-tailed unpaired t-test, $p < 0.05$).
    }
    \label{tab:results}
    \vspace{-4mm}

\end{table}

\begin{table*}[t]
\centering
    \include{tables/breakdown}
    \caption{Breakdown of F1 performance per example split, with each split comprising reviews containing that quadruple type. E: explicit, I: implicit, A: aspect, O: opinion. 
    $\dagger$: method is significantly better than PARAPHRASE (one-tailed unpaired t-test, $p < 0.05$).
    }
    \label{tab:results_breakdown}
    \vspace{-5mm}
\end{table*}

\begin{table}[t]
\centering
    \include{tables/ablation}

    \caption{Ablation analysis of the GEN-SCL-NAT model. F1 scores are reported, averaged over 5 runs. We compare to the baseline PARAPHRASE model. The best result is bolded. $\dagger$: method is significantly better than PARAPHRASE (one-tailed unpaired t-test, $p < 0.05$).}
    \label{tab:ablation}
    \vspace{-5mm}
\end{table}

\vspace{-2mm}
\paragraph{Model Comparisons} We compare five ACOS techniques, considering both implicit aspect and implicit opinion cases as done in \citet{acos}.

$\bullet$ \textbf{TAS-BERT-ACOS}: \citet{acos} devise a two-step pipelined method, incorporating TAS-BERT \cite{wan2020target} for triplet extraction.

$\bullet$ \textbf{Extract-Classify-ACOS}: \citet{acos} leverage BERT \cite{devlin2018bert} to extract aspect-opinion pairs then perform category and sentiment linking.

$\bullet$ \textbf{Seq2Path}: \citet{seq2path} generate ACOS quadruples as paths of a tree, supporting multiple quadruples through multi-beam search and filtering candidates via a learned discriminator token.

$\bullet$ \textbf{PARAPHRASE}: A generative T5 structured paraphrase generation model, producing a parseable ACOS sequence prediction~\cite{asqp}.

$\bullet$ \textbf{GEN-SCL-NAT}: Finally, our proposed method, including the SCL and NAT components.

\paragraph{Experiment Details} We report scores averaged over five runs each with different random seeds. For our GEN-* models, we adopt the 770M parameter T5-large \cite{t5} as our pre-trained generative encoder-decoder model. We report further experiment details and hyperparameter settings in Appendix \ref{sec:appendix_experiment_details}.
\section{Results}
\label{section:results}
We evaluate on the task of exact quadruple extraction using the F1 metric, where a correct extraction requires all components are correct.

\paragraph{Overall Performance}
Table \ref{tab:results} reports the overall performance on the ACOS task.
We find that GEN-SCL-NAT outperforms other approaches on all three datasets. The BERT pipeline approaches struggle heavily, likely due to error-accumulation over several sub-tasks and limited pre-training alignment. 
Among T5 methods, Seq2Path performs worst by over 3\% on the two reported datasets, perhaps due to limitations in their beam-search candidate pruning method; in contrast PARAPHRASE and our approach generate all quadruples in one output sequence.

\paragraph{Explicit vs Implicit Performance Breakdown}
Table \ref{tab:results_breakdown} reports the results on four implicit/explicit aspect/opinion data splits. Notably, our method is adept at the IAEO and EAIO subsets, with average 2.56\% and 1.37\% respective increases of F1 over PARAPHRASE. We see *IO splits are consistently more challenging than *EO splits, as models have no supervised signal with which to localize implicit opinions in the text. However, we still see consistent EAIO gains from our method, along with IAIO gains of 1.5\% for the REST dataset. These results speak to the improved predictive power of our techniques---we consistently outperform on the difficult and sparse implicit subsets while still increasing the average EAEO performance by 0.88\%.

\section{Additional Analyses}

\subsection{GEN-SCL-NAT Ablation Study}

We ablate the GEN-SCL-NAT model by withholding various components of the technique, including each of the NAT enhancements and SCL losses. Table \ref{tab:ablation} reports ablation results.

\paragraph{SCL Ablation}
\vspace{-1.5mm}
We ablate each of the three SCL objectives.
We withhold each component of the SCL loss, finding the opinion loss most impactful (average decrease of 1.03\%), and also observe decreases from the aspect and sentiment ablations. Notably, we see the losses considering implicit language phenomena (\textit{Aspect SCL}, \textit{Opinion SCL}) are the most impactful, indicating the benefits of modeling these challenging examples. We additionally withhold all SCL losses (\textit{All SCL}), observing noticeable consistent performance degradation on all datasets with the SCL component removed.

\begin{figure*}
\vspace{-4mm}
\centering
\captionsetup[subfigure]{labelformat=empty}
\centering
\begin{subfigure}{.32\textwidth}
\centering
\includegraphics[height=4.6cm]{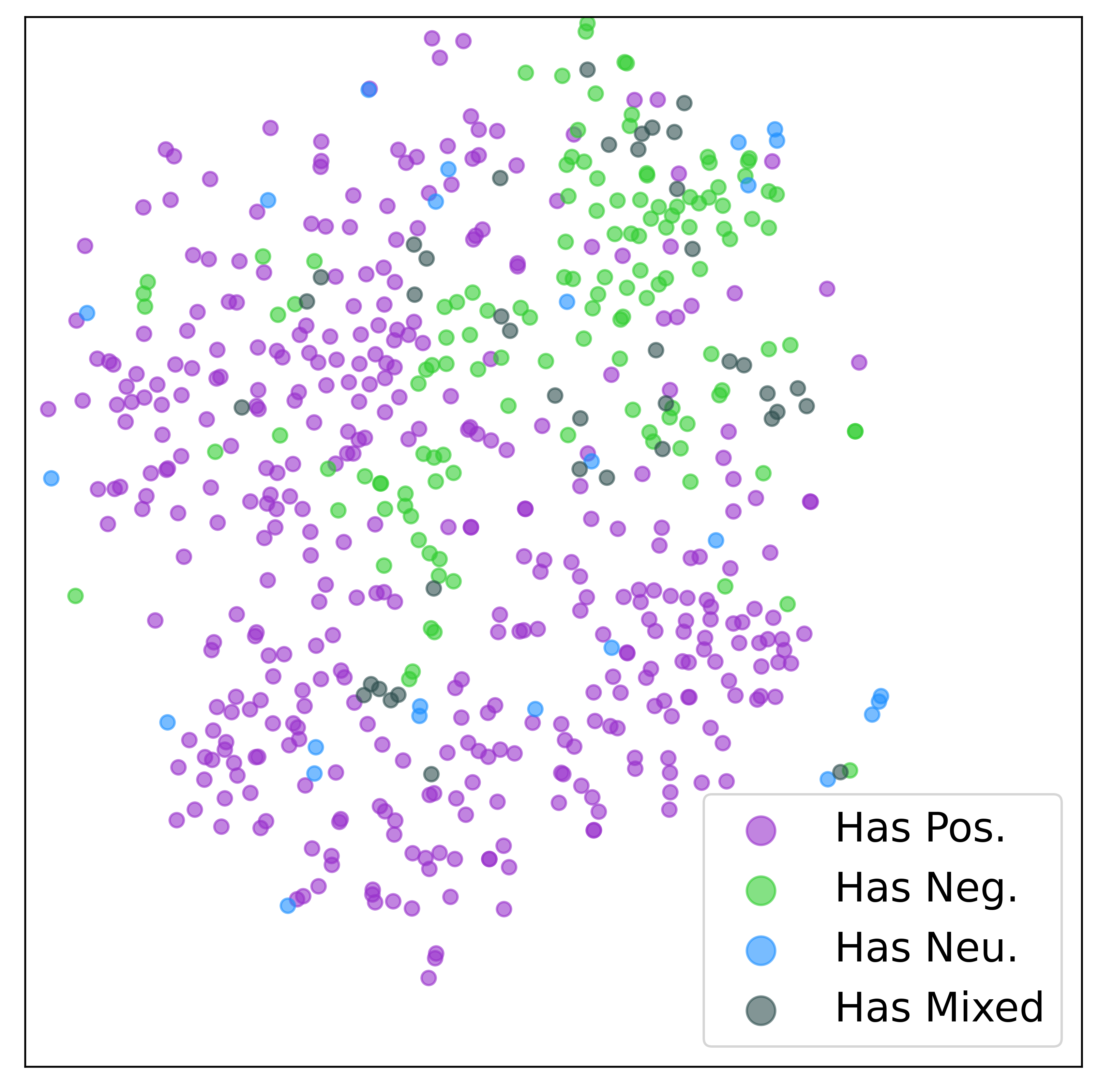}
\caption{Sentiment w/o GEN-SCL}
\end{subfigure} 
\begin{subfigure}{.32\textwidth}
\centering
\includegraphics[height=4.6cm]{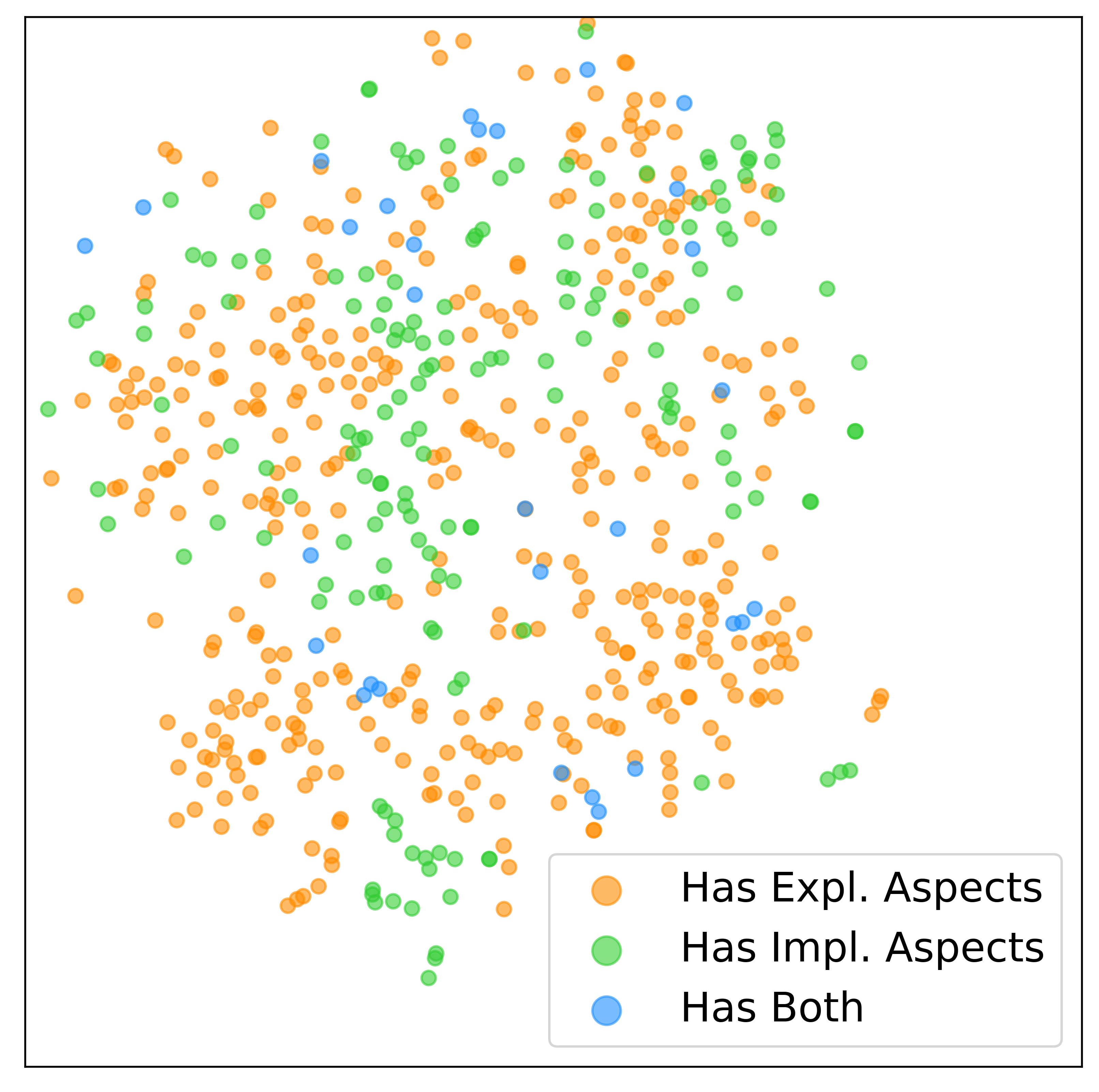}
\caption{Aspect w/o GEN-SCL}
\end{subfigure} 
\begin{subfigure}{.32\textwidth}
\centering
\includegraphics[height=4.6cm]{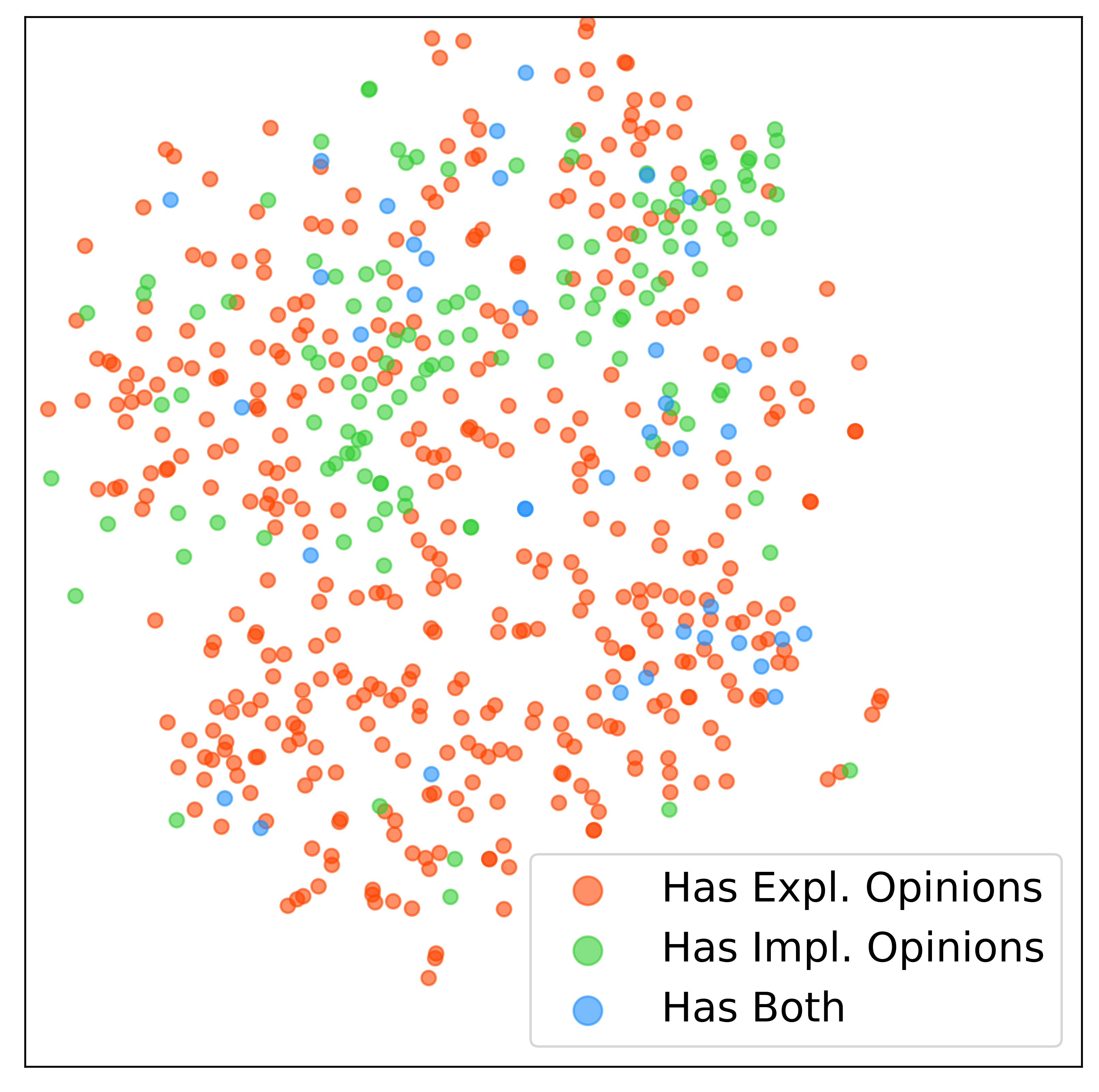}
\caption{Opinion w/o GEN-SCL}
\end{subfigure} \\
\begin{subfigure}{.327\textwidth}
\centering
\includegraphics[height=4.6cm]{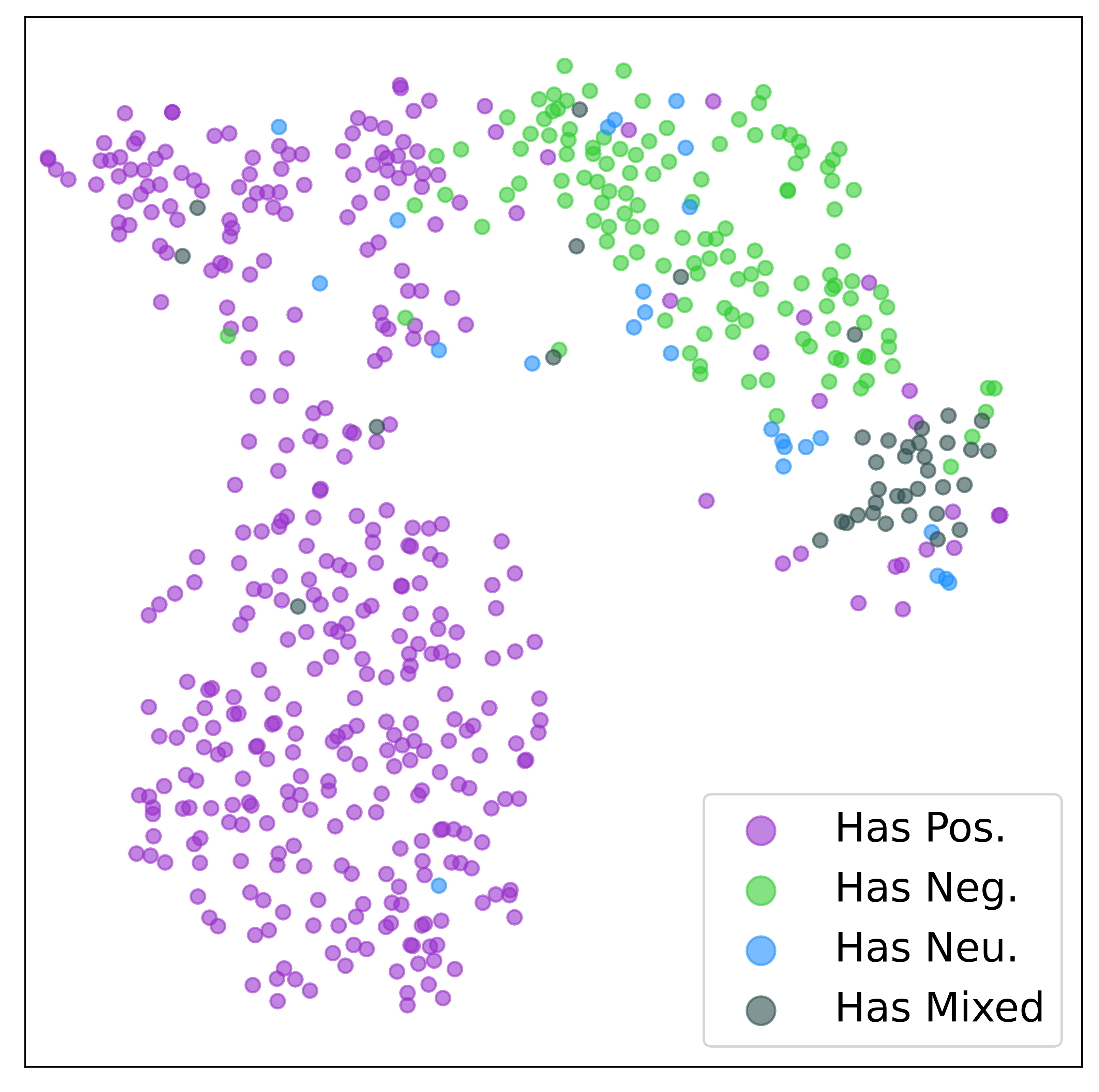}
\caption{Sentiment w/ GEN-SCL}
\end{subfigure}%
\begin{subfigure}{.327\textwidth}
\centering
\includegraphics[height=4.6cm]{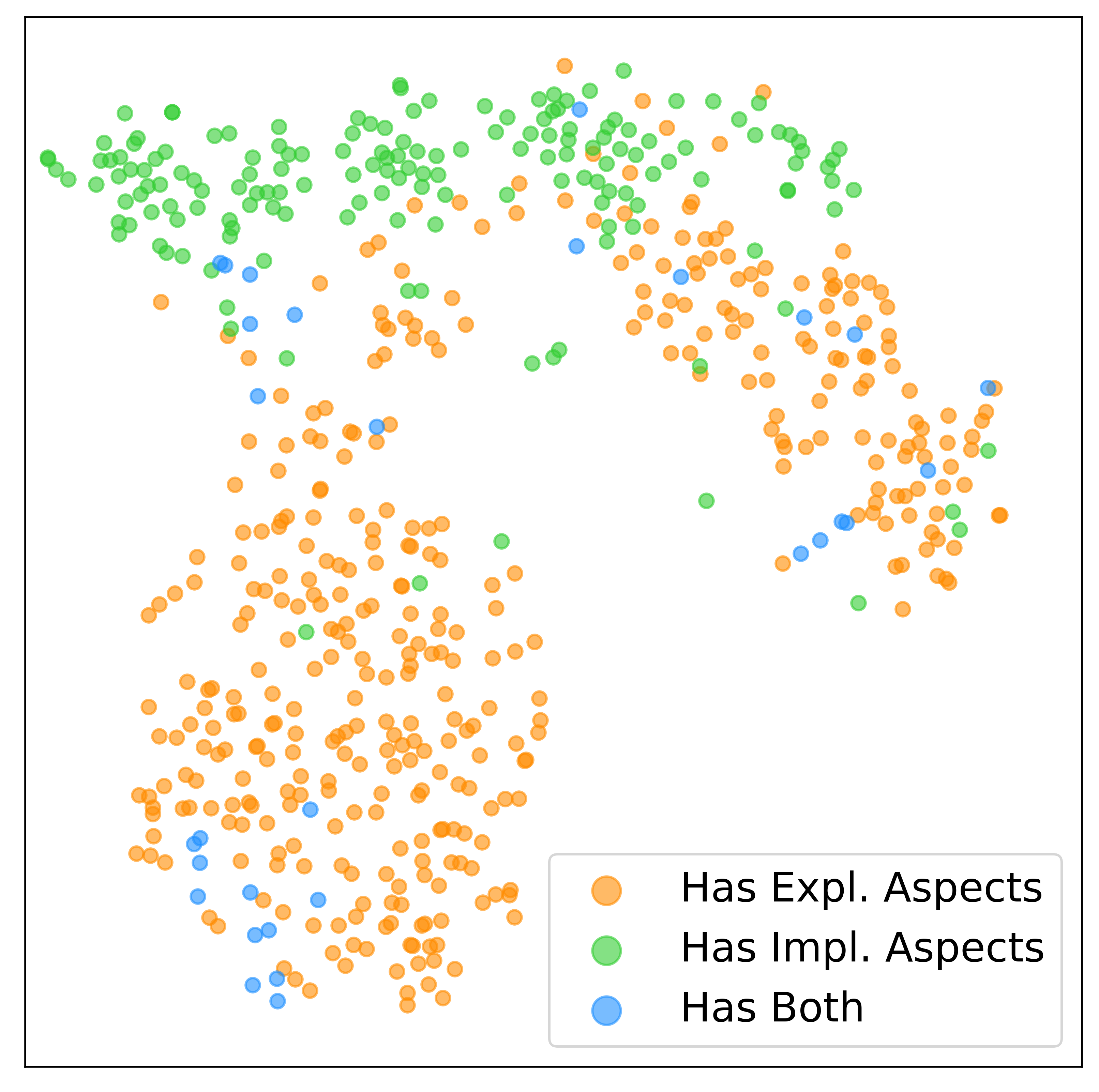}
\caption{Aspect w/ GEN-SCL}
\end{subfigure}%
\begin{subfigure}{.327\textwidth}
\centering
\includegraphics[height=4.6cm]{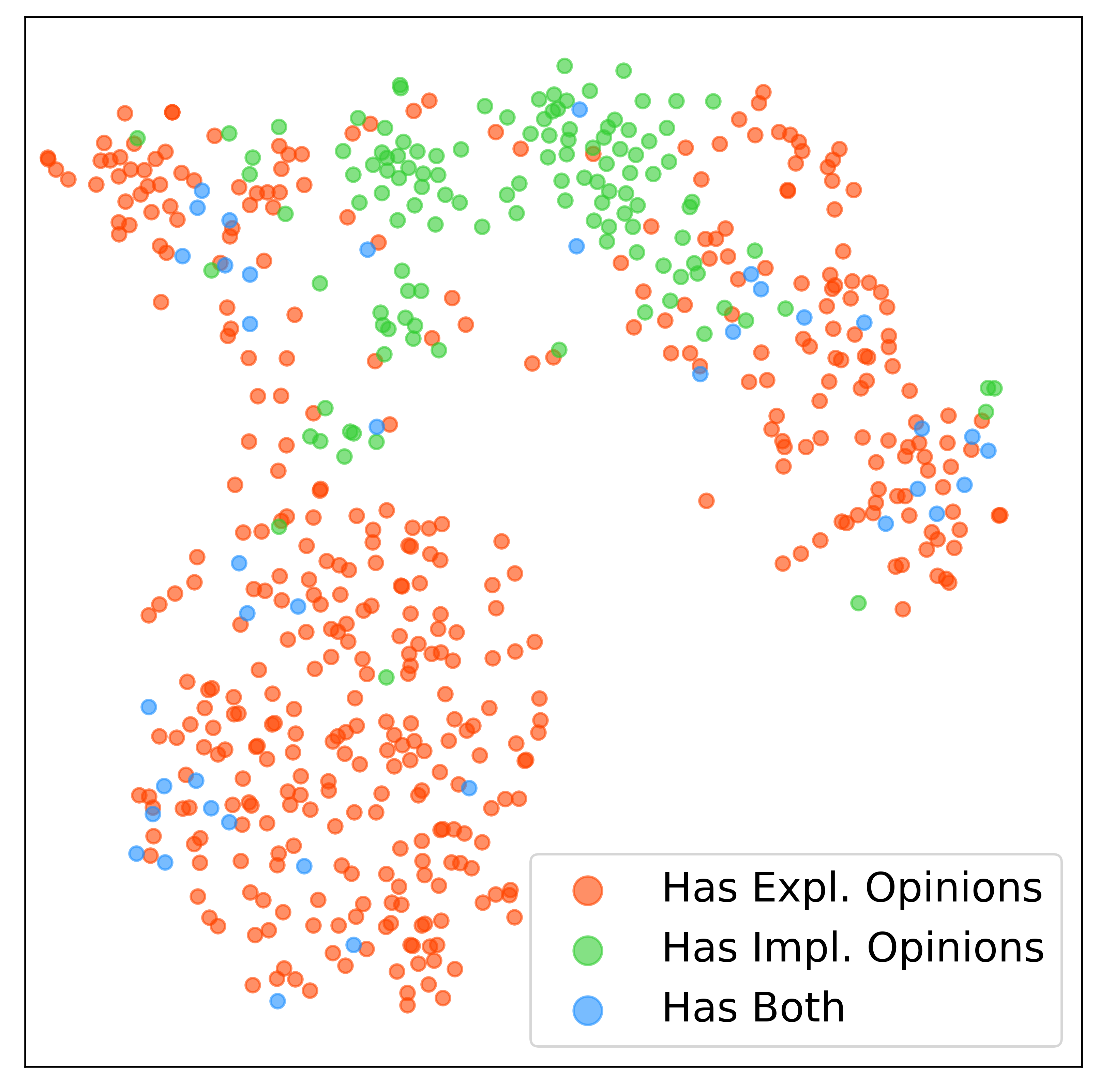}
\caption{Opinion w/ GEN-SCL}
\end{subfigure} 

\caption{
T-SNE visualization of mean-pooled encoder final layer on the Restaurant dataset. 
Our GEN-SCL objective encourages the encoder to produce with discriminable representations of three key input characteristics. 
}
\label{fig:tsne}
\vspace{-4mm}
\end{figure*}

\paragraph{NAT Ablation}
\vspace{-1.5mm}
Of the three NAT components, we see that excluding scan-based multi-quad ordering (outputting quadruples in the order they are mentioned in the text) has the largest impact on overall performance. Notably, the impact is largest on the REST dataset $(62.62 \rightarrow 59.92)$ which contains more quadruples per example than the Laptop datasets (average of 1.60 vs. 1.42). This validates our intuition that proper output ordering is significant for multi-quadruple examples.
Next, we see that natural category labels are impactful, especially for the LAPTOP dataset $(45.16 \rightarrow 43.94)$ containing a much larger category labelset (121 labels versus 13 and 21 for the Restaurant and Laptop-L1 datasets). 
Finally, we see GEN-SCL-NAT benefits from the intra-quad ordering, although not as noticeably as the other two improvements. We see improvements on REST and LAPTOP-L1, while performance was comparable with and without for LAPTOP $(45.16 \rightarrow 45.43)$. 
Overall, our full GEN-SCL-NAT method yields consistently strong performance, with its components working in unison to address challenging properties that arise in the ACOS task (implicit aspects and opinions, multi-quad examples, large category labelsets). 

\subsection{GEN-SCL Representations}
To better understand the behavior of the GEN-SCL objective on the model hidden representations, we generate t-SNE \cite{tsne} visualizations of the mean-pooled final encoder layer. Figure \ref{fig:tsne} displays results on the Restaurant test set. We see GEN-SCL enables the encoder to simultaneously represent sentiment, aspect, and opinion information effectively.
\vspace{-2mm}
\color{black}

\section{Related Work}
\vspace{-2mm}
ACOS quadruple prediction is an emerging ABSA task with structured generation techniques producing top results on the ACOS task \cite{asqp, seq2path, surveyzhang2022}. \citet{keyphrase-gen} empirically demonstrate the importance of output ordering and formatting in generative unordered set prediction techniques for keyphrase generation, motivating our exploration of this direction.
Supervised contrastive learning \cite{supcon_loss} is a popular technique for representation learning and has proven useful for NLP tasks \cite{supcl-seq}. \citet{scaptABSA} apply it to ABSA tasks during pre-training, but only for representing positive vs negative sentiment. Our novel methods combine generation format improvements with a task-specific supervised contrastive learning objective that learns to represent key ACOS characteristics.

\section{Conclusion}
\vspace{-2mm}
In this work, we introduce GEN-NAT, a modified ACOS generation output format encompassing three methods for improving the naturalness of the decoded output sequences. We combine this with GEN-SCL, our novel task-specific application of supervised contrastive learning that learns improved example representations leading to downstream gains. Our proposed GEN-SCL-NAT method demonstrates state-of-the-art results, both overall and for challenging implicit-sentiment splits.

\section*{Acknowledgements}
This work is supported in part by the National Science Foundation through grant CMMI-2050130. We thank the anonymous reviewers for their valuable feedback and suggestions. 

\color{black}

\section*{Limitations}
While effective on concise sentence-level tasks, we have not yet explored ACOS structured generation for document-level and/or sentiment-dense inputs. Approaches such as ours and \newcite{asqp} generate the prediction as a single output, and we may encounter issues such as output structure validity when handling longer examples with higher quadruple frequency.
Our evaluation of GEN-SCL is also confined to generative sequence prediction models. While it may generalize to other formulations, we have not yet explored this direction.

\bibliographystyle{conf/acl_natbib}
\bibliography{emnlp22}

\appendix
\section{Implementation Details}
\label{sec:appendix_experiment_details}
We use Huggingface Transformers \cite{huggingface} for training our T5 models. We conduct our experiments on the Nvidia A40 GPU with 48GB VRAM.
\paragraph{Training}
We largely use the T5 parameters specified in \citet{asqp}, making the following adjustments after performing manual accuracy-based hyperparameter tuning on the provided validation sets: We train for 45 epochs using batch size of 32, and a learning rate of 9e-5 for all experiments. For GEN-SCL, we tune $\alpha$, the SCL loss weighting factor, and the $\tau$ temperature parameter per dataset, with these values reported in Table \ref{tab:scl_params}. 
We use the standard dataset train/validation/test splits provided by the authors of the REST and LAPTOP* datasets. Training takes approximately 1 hour for the RESTAURANT dataset, and 1.5 hours for the LAPTOP* datasets.
\paragraph{Prediction}
We use beam search to decode the ACOS output sequence; we set beam size to 5. Prediction on the RESTAURANT and LAPTOP test sets takes 5 minutes with an evaluation batch size of 32.

\begin{table}[]
\centering
\include{tables/scl_params}
\caption{We report the parameters used in the GEN-SCL supervised contrastive learning objective. $\alpha$ is the loss weighting factor, and $\tau$ is the temperature value that determines how severely to punish hard negative examples.}
\label{tab:scl_params}
\end{table}

\begin{table}[]
\centering
\include{tables/category_examples}
\caption{Examples of the aspect categories used for the ACOS task. Our GEN-NAT technique better leverages T5 pre-training by using human-readable descriptive category labels.}
\label{tab:category_mappings}
\end{table}

\section{GEN-NAT Category Mappings}
One component of our GEN-NAT approach consists of replacing existing raw aspect category labels with human-readable natural category descriptions. Table \ref{tab:category_mappings} provides examples of these reformatted GEN-NAT category labels.

\end{document}

%% file: tables/dataset_stats.tex
\resizebox{\columnwidth}{!}{%
\begin{tabular}{|l|l|l|l|}
\hline
                    & REST & LAPTOP   & LAPTOP-L1 \\ \hline
\#Categories        & 13              & 121           & 21                    \\ \hline
\#Sentences         & 2286            & 4076          & 4076                  \\ \hline
\#EAEO Quads            & 2429 (66.4\%)   & 3269 (56.8\%) & 3269 (56.8\%)         \\ \hline
\#IAEO Quads          & 530 (14.5\%)    & 910 (15.8\%)  & 910 (15.8\%)          \\ \hline
\#EAIO Quads           & 350 (9.57\%)    & 1237 (21.5\%) & 1237 (21.5\%)         \\ \hline
\#IAIO Quads         & 349 (9.54\%)    & 342 (5.94\%)  & 342 (5.94\%)          \\ \hline
\#Quads/Sent & 1.60            & 1.42          & 1.42                  \\ \hline
\end{tabular}
}

%% file: tables/primary_results.tex
\resizebox{\columnwidth}{!}{%
\begin{tabular}{llll}
\hline
                      & \multicolumn{1}{c}{REST}                 & \multicolumn{1}{c}{LAPTOP}  & \multicolumn{1}{c}{LAPTOP-L1}          
             \\ \hline
\textit{BERT Backbone} \\
  TAS-BERT-ACOS$^*$            & 33.53                 & 27.31             & --          \\
  Extract-Classify-ACOS$^*$          & 44.61              & 35.80              & --                \\ \hline
\textit{T5 Backbone} \\
Seq2Path$^*$      & 58.06                  & 41.45               & --   \\ 
PARAPHRASE                      & 60.97                & 44.08           & 60.73         \\ 
GEN-NAT-SCL (ours)   & \textbf{62.62$^\dagger$}  & \textbf{45.16$^\dagger$}  & \textbf{62.46$^\dagger$}         \\
\hline

\end{tabular}%
}

%% file: tables/breakdown.tex
\resizebox{\textwidth}{!}{%
\begin{tabular}{lllllllllllll}
\hline
                      & \multicolumn{4}{c}{REST}                                                    & \multicolumn{4}{c}{LAPTOP}                                                        & \multicolumn{4}{c}{LAPTOP-L1}                              \\ \cline{2-13} 
Method                & EAEO           & IAEO           & EAIO           & \multicolumn{1}{c|}{IAIO}           & EAEO           & IAEO           & EAIO           & \multicolumn{1}{c|}{IAIO}           & EAEO           & IAEO           & EAIO           & IAIO           \\ \hline
TAS-BERT-ACOS         & 33.6          & 31.8          & 14.0          & \multicolumn{1}{c|}{39.8}          & 26.1          & 41.5          & 10.9          & \multicolumn{1}{c|}{21.2}          & -            & -            & -            & -            \\
Extract-Classify-ACOS & 45.0          & 34.7          & 23.9          & \multicolumn{1}{c|}{33.7}          & 35.4          & 39.0          & 16.8          & \multicolumn{1}{c|}{18.6}          & -            & -            & -            & -            \\ 
PARAPHRASE             & 65.4          & 53.3          & 45.6          & \multicolumn{1}{c|}{49.2}          & 45.7          & 51.0          & 33.0          & \multicolumn{1}{c|}{\textbf{39.6}}          & 64.2          & 65.2          & 49.3          & \textbf{53.8} \\ 
GEN-SCL-NAT (ours)       & \textbf{66.5}         & \textbf{56.5$^\dagger$}          & \textbf{46.2} & \multicolumn{1}{c|}{\textbf{50.7}} & \textbf{45.8}          & \textbf{54.0$^\dagger$}          & \textbf{34.3}          & \multicolumn{1}{c|}{\textbf{39.6}}          & \textbf{65.6}          & \textbf{66.7} & \textbf{51.5$^\dagger$} & 53.7\\
\hline
\end{tabular}
}

%% file: tables/ablation.tex
\resizebox{\columnwidth}{!}{%
\begin{tabular}{llll|c}
\hline
               & REST & LAPTOP & \multicolumn{1}{c}{LAPTOP-L1} & \multicolumn{1}{c}{Avg. $\Delta$} \\ \hline
PARAPHRASE     & 60.97           & 44.08       & 60.73 & -1.49 \\
GEN-SCL-NAT    & \textbf{62.62$^\dagger$}           & 45.16$^\dagger$       & \textbf{62.46$^\dagger$}  & -                         \\
\midrule
\hspace*{1mm}(-Sentiment SCL) & 62.36$^\dagger$           & \textbf{45.72$^\dagger$}           & 62.03$^\dagger$  & -0.05                                \\
\hspace*{1mm}(-Aspect SCL)    & 61.78           & 45.14$^\dagger$       & 61.68$^\dagger$       & -0.55                  \\
\hspace*{1mm}(-Opinion SCL)   & 60.68           & 44.71       & 61.77$^\dagger$    & -1.03                              \\
\hspace*{1mm}(-All SCL)       & 62.18$^\dagger$           & 44.03       & 62.12$^\dagger$  & -0.64                            \\
\midrule
\hspace*{1mm}(-Multi-quad Ordering)    & 59.92          & 44.63      & 61.00     & -1.56                  \\
\hspace*{1mm}(-Natural Category Labels)   & 61.28           & 43.94       & 61.96$^\dagger$    & -1.02                              \\
\hspace*{1mm}(-Intra-quad Ordering)       & 61.39          & 45.43$^\dagger$       & 61.48$^\dagger$  & -0.65                            \\
\hspace*{1mm}(-All NAT)           &   59.89              &  43.86           & 61.36 &  -1.26  \\

\hline
\end{tabular}
}

%% file: tables/scl_params.tex
\begin{tabular}{lcc}
\hline
          & $\alpha$ & $\tau$ \\ \hline
REST      & 0.05                  & 0.25                \\
LAPTOP    & 0.05                  & 0.25                \\
LAPTOP-L1 & 0.005                 & 0.25           \\
\hline
\end{tabular}

%% file: tables/category_examples.tex
\resizebox{\columnwidth}{!}{%
\begin{tabular}{lcc}
\hline
Dataset   & Raw Category Label                  & GEN-NAT Category Label                       \\ \hline
REST      & LOCATION\#GENERAL    & the location                  \\
          & FOOD\#PRICES         & the food prices               \\
          & FOOD\#QUALITY        & the food quality              \\ \midrule
LAPTOP    & OS\#GENERAL          & the operating system overall  \\
          & OS\#DESIGN\_FEATURES & the operating system features \\
          & HARD\_DISC\#PRICE    & the hard drive price          \\ \midrule
LAPTOP-L1 & OS                   & the operating system          \\
          & HARD\_DISC           & the hard drive                \\ \hline
\end{tabular}
}